\documentclass[a4paper,10pt]{amsart}
\usepackage[utf8]{inputenc}
\usepackage{amsmath, amsthm, amscd, amsfonts, amssymb, graphicx, color}
\usepackage[bookmarksnumbered, colorlinks, plainpages]{hyperref}
\usepackage{wrapfig}
\usepackage{subfig}
\usepackage{color,graphicx,enumerate,amssymb}
\usepackage{hyperref}
\usepackage{graphicx}
\usepackage[export]{adjustbox}
\usepackage{eucal}
\usepackage{algorithm}
\usepackage{algpseudocode}
\usepackage[utf8]{inputenc}
\usepackage{dsfont}
\usepackage{mathtools}
\usepackage{comment}
\usepackage{bbm}

\usepackage{type1cm}   
\usepackage{lineno}

\newcommand{\bR}{{\mathbb R}}

\newcommand{\cL}{{\mathcal L}}

\newcommand{\cX}{{\mathcal X}}

\newcommand{\uu}{\mathbf{ u}}

\newcommand{\rdiv}{{\rm div\,}}

\newtheorem{theorem}{Theorem}[section]

\newtheorem{example}[theorem]{Example}

\title [ Improved Graph- based semi-supervised learning Schemes ]{ Improved Graph- based semi-supervised learning Schemes }
\author{Farid Bozorgnia}

\address{CAMGSD, Department of Mathematics, Instituto Superior T\'{e}cnico, Lisbon, Portugal}
\email{faridb.bozorgnia@tecnico.ulisboa.pt}

\thanks{F. Bozorgnia was supported by the FCT  fellowship SFRH/BPD/33962/2009 and by Marie Skłodowska-Curie grant agreement No. 777826 (NoMADS)}

\subjclass[2000]{}


\begin{document}
\begin{abstract}
In this work,  we  improve the accuracy  of     several known  algorithms  to address the  classification of large datasets when few labels are available. Our framework lies in the realm of  graph-based semi-supervised learning. With novel modifications  on  Gaussian Random Fields Learning and   Poisson Learning  algorithms, we increase the accuracy and  create   more  robust algorithms. Experimental results demonstrate the efficiency  and superiority of the proposed methods over conventional graph-based semi-supervised  techniques, especially in the context of imbalanced datasets.
 
\end{abstract}
\bigskip

\maketitle

\section{Introduction}

In this paper, we consider  semi-supervised learning approaches   for    classifying   many  unlabeled data when very few labels   are  given.    Our objective is to refine existing classification algorithms to perform effectively in the cases with    limited labeled data. We introduce innovative modifications to established algorithms to enhance accuracy and robustness.

 When there is an insufficient amount of labeled data available, models may struggle to learn effectively and to predict accurately. One way to improve this is by  incorporating lots of unlabeled data along with  those few  labeled samples. This approach helps the model to  learn  about the structure of   data, enhancing  its performance especially when there aren't many examples to learn from. Considering  mentioned advantage,    semi-supervised learning has received much attention in the past few decades \cite{SG, CSZ, Asp, A20}. A  common approach   for doing this is  graph-based semi-supervised learning.  We use graphs to show how different samples  of data are connected, helping to understand both labeled and unlabeled data better. We refer to  recent surveys  \cite{SYXK, CDYS} as a comprehensive review of existing algorithms  on graph-based learning.

Although existing semi-supervised algorithms   show good performance on balanced data, they frequently struggle in complex real-world applications, particularly with highly imbalanced datasets. One of the  issues is overfilling to the limited labeled data available and not making full use of the structural information present in the unlabeled data. The  Imbalance Ratio (\textit{IR}) is an important factor  and  represents the ratio of samples in each of \textit{Minority} and \textit{Majority} classes.  The bigger \textit{IR}, the more difficult the problem becomes, to see more about techniques  to  handle  imbalanced data see \cite{Rasool}.  Our objective is to introduce innovative modifications to established algorithms, enhancing both accuracy and robustness against these challenges.

A common approach   to use unlabeled data in semi-supervised learning is to build a graph over the data. This means  geometric or topological properties of the unlabeled data have been used  to   improve  algorithms.  In \cite{Liang}   a general framework for constructing a  graph is proposed and they named it    Graph-based Semi-supervised Learning via Improving the Quality of the Graph Dynamically.

First, we   construct  a  weight matrix  $W$ or  similarity  matrix  over the data set, which encodes the similarities between pairs of data nodes. If our data set consists of $n$ points  $\cX=\{x_1,  x_2, \cdots ,x_n \}  \subset \mathbb{R}^d$, then the  weight matrix $W$   is an  $n \times n $  symmetric matrix, where the element  $w_{ij}$  represents the similarity between two data points  $x_i$  and $x_j$. When constructing a KNN graph, the weight   between node $x_i$  and node $x_j$  typically depends on the distance or similarity between the nodes. A  common choice for the weight function is  the Gaussian kernel or the inverse of the distance between nodes.

The similarity is always nonnegative and should be large when  $x_i$  and $x_j$ are close together spatially,
and small (or zero), when $x_i$  and $x_j$  are far apart.  Subsequently, by leveraging information from the constructed graph, labels are propagated from the given labeled instances to the unlabeled data.

The main contribution of the  current work is as follows. 
\begin{itemize}
\item   Modified GRF and Improved GRF Learning Algorithms: To speed up convergence and handle imbalances more effectively by incorporating the stationary distribution $\pi$ of the random walk on the graph.
    
\item Improved Poisson Learning Algorithm: Introduces regularization terms to enhance performance on imbalanced data.
\end{itemize}

\section{Problem   setting  and previous works }

This section is devoted as an introduction to  PDEs  on  graph  and setting our problem.
Let $\cX=\{x_1,\cdots,x_n\}$ denote the vertices of a  graph with   symmetric edge weight $w_{ij}$ between $x_i,x_j\in \cX$ and let $W=[w_{ij}]$. We assume there is a subset of the nodes $ \cX_{l} ={\{ x_1, \cdots, x_l}\}  \subset  \cX $ for which labels are provided by a labeling function  $ g: \cX_{l}\rightarrow\bR^k$, forming our training set    $(x_i, y_i)_{i=1}^{l}$.
The degree of a vertex $x_i$ is given by $d_i=\sum_{j} w_{ij}$ and let $d=\sum_{i} d_i$.  Matrix $D$ denotes the diagonal matrix which has $d_i$ on main the diagonal. 

 Let $\ell^{2}(\cX)$ denote the set of functions $u: \cX \rightarrow  \mathbb{R}$ equipped with the inner product
\[
(u,v)=\sum\limits_{x\in \cX}  u(x)v(x),
\]
for functions $ u,v : \cX\rightarrow \mathbb{R}$.
We also define a vector field on the graph to be an antisymmetric function $V:\cX\times \cX\rightarrow \mathbb{R}^2$, i.e. $V(x,y)=-V(y,x)$ and denote the space of all vector fields by $\ell^2(\cX^2)$.

The gradient of a function $ u\in \ell^{2}(\cX) $ is the vector field
\[
\nabla u(x,y)=u(y)-u(x).
\]
The unnormalized graph laplacian $\cL $ of a function $u\in \ell^2(\cX)$ is defined as
\[
\cL u(x):=-\rdiv(\nabla u)(x)=\sum_{y\in \cX}w_{xy}(u(x)-u(y)).
\]
Most graph-based semi-supervised learning aims to find a function on the graph that closely matches the given labels while also maintaining smoothness. It is shown in \cite{A21} that the graph Laplacian regularization is effective because it  forces  the labels to be consistent with the graph structure.

To model label propagation in semi-supervised learning, it is assumed that the learned labels vary smoothly and do not change rapidly within high-density regions of the graph (smoothness assumptions) \cite{BNS}. 
Based on this assumption different approaches have  been proposed, we refer to  the pioneer methods   {\it Laplace learning}, \cite{A19} see also \cite{A3, BNS}.

Let $U$ stand for  the   label matrix $U$  that shows  the class information
of each element.  A  general form of a graph-based
semi-supervised learning  for data classification can be formulated  as a 
minimization of an energy:
\[
R(U) +  \mu F(U),
\]
where  $R(U)$  is a regularization term
incorporating the graph weights, and $F(U)$  is a forcing term,
which usually incorporates the labeled points and their class
information.

 In Laplace learning algorithm the labels are extended by finding the minimizer   $ U: \cX\rightarrow \mathbb{R}^k$ for the following problem
 \begin{equation}\label{objective}
 \begin{cases}
&\min J_n(U):=\frac{1}{2}\sum\limits_{i=1}^{n}\sum\limits_{j=1}^{n} w_{ij} |U(x_{i}) - U(x_{j})|^{2}\\
&\text{subject to } U(x_i) =y_i,\quad  \text{ for } i=1, 2, \cdots ,l.
\end{cases}
\end{equation}
The minimizer  will be a harmonic function satisfying
\begin{equation*}
\left \{
\begin{array}{ll}
\cL U(x)=0,  &  x\in \cX\setminus \cX_{l},\\
  U=g,   &   \text{ on } \cX_{l},\\
  \end{array}
\right.
\end{equation*}
 where $\cL$ is the unnormalized graph Laplacian given by
  \begin{equation*}
 \mathcal{L}U(x_i)=\sum\limits_{j=1}^{n} w_{ij}\, (U(x_{i}) - U(x_j)).
 \end{equation*}
Let $U=(u_1,\cdots,u_k)$ be a solution of \eqref{objective}, the label of node $x_i\in \cX\setminus\cX_{l}$ is dictated by
 \begin{equation*}
\underset{j\in {\{1,\cdots, k}\}}{ \textrm{arg max}} {u_{j}(x_i)}.
  \end{equation*}
 This means that  Laplace  learning uses harmonic extension on a graph to
propagate labels. If  the number of labeled data samples is finite
while the number of unlabeled data   tends to infinity, then Laplace learning
becomes degenerate and the solutions become roughly constant with a spike at each
labeled data point.

Later it has been observed that the Laplace learning can give poor results in classification \cite{A18}.
The results are often poor because the solutions have localized spikes near the labeled points, while being almost constant far from them.
To overcome this problem several versions of the  Laplace learning algorithm  have been  proposed, for instance, Laplacian regularization,  \cite{A1}, weighted Laplacian, \cite{ A14} and $p$-Laplace learning, \cite {A10, slepcev2019analysis}.
Also, the limiting  case in $p$-Laplacian when $p$ tends to infinity  is so-called Lipschitz learning is studied in \cite{K17} and similar to continuum PDEs  is related to finding  the absolute  minimal Lipschitz extension of the training data.
Recently,   in \cite{A8}  another approach to increase   accuracy of Laplace learning  is   given  and called {\it Poisson learning}.

In \cite{A12} the authors  consider the case that   the number of labeled data points  grows to
infinity  also when  the total number of data points grows. Let  $\beta$  denote the labeling rate.  They   show that for a random geometric
graph with length scale  $\varepsilon$  if  $\beta \ge \varepsilon^{2}$, then the solution becomes
degenerate and spikes occurs, while   for the case $\beta \le \varepsilon^{2}$,  Laplacian learning is
well-posed and consistent with a continuum Laplace equation.

The authors in \cite{A8}  have proposed a  scheme, called Poisson learning that  replaces the  label values at training points  as sources and sinks, and solves the Poisson equation on the graph as follows:
\begin{equation}\label{eq:state13}
\left \{
\begin{array}{ll}
\mathcal{L}\uu(x_i) = y_{i}- \overline{y}   &  1\le i\le l,\\
  \mathcal{L}\uu(x_i)=0  &  l+1\le i\le n,\ \\
  \end{array}
\right.
\end{equation}
with further condition    $\sum_{i=1}^{n} d(x_i) \uu(x_i)  =0$, where
$
\overline{y}=\frac{1}{l}\sum_{i=1}^{l} y_{i}
$
is the average label vector. In the next section, we review this scheme in detail and will consider some modification to this algorithm. 

In \cite{AB, ABR} a graph-based semi-supervised learning scheme based on  the theory of the spatial segregation of competitive systems, is given. The scheme is based on the minimizing  \eqref{objective} with   constraint
\begin{equation*}
u_i(x)\cdot u_j(x)=0, \qquad\text{ for all }x\in \cX, \;1\leq i\neq j\leq k.
\end{equation*}

\section{Proposed Algorithms }

\subsection{Modified   Gaussian Random Fields (MGRF) }
 Let,  as before  $\cX={\{ x_1, \cdots, x_n}\}$ denote the data points or vertices in a graph.
  We assume there is a subset of the nodes $ \cX_{l} ={\{ x_1, \cdots, x_l}\}  \subset  \cX$  that their labels are given with  a label function $g: \cX_{l}\rightarrow\bR^k$.
It is further assumed that $y_i=g(x_i)\in \{e_1, \cdots , e_k\}$ where $e_i$ is the standard basis in $\mathbb{R}^k$ and represents the $i\textsuperscript{th}$ class.
In graph-based semi-supervised learning, we aim to  extend labels  to the rest of the vertices
 $\cX_{u}= \{x_{l+1}, \cdots, x_n\}$.

The scheme described  below is known as Gaussian Random Fields (GRF)   \cite{A19}.  For any matrix  $U=(U_{ij})\in \mathbb{R}^{n\times k}$, $U_{i\cdot}$  represents  the  $i^{\text{th}}$  row of $U$. They consider the following minimization problem
\begin{equation}\label{GRF}
 \underset{U_{l\cdot}=Y_{l\cdot}}{\text{arg min}} \, Tr(U^T L U), \quad U \in \mathbb{R}^{n\times k}
 \end{equation}
 Here $L=D-W$.  Let us split the weight matrix $W$ into four blocks as 
\begin{displaymath}
\mathbf{W}=
\left(\begin{array}{cc}
W_{ll} & W_{lu} \\
W_{ul}  &  W_{uu}\\
\end{array}\right).
\end{displaymath}
Also let $P=D^{-1} W$  and  decompose $P$ and  $U$ to $U_l$ and $U_u$,  then the  solution of minimization problem (\ref{GRF}) is  given by:

\begin{equation}\label{Sun11}
  U_{u} =(I -P_{uu})^{-1} P_{ul} U_l.
  \end{equation}
The  following iterative scheme was proposed in  \cite{A19}.
\begin{equation}\label{GRF22}
 U^{(m+1)}=   \alpha S U^{(m)}  +(1-\alpha) Y,
\end{equation}
  where   $S= D^{\frac{-1}{2}}  W  D^{\frac{-1}{2}}$ and  $Y$  denotes  the vector  of   initial labels.

We modify the scheme  (\refeq{GRF22}) to achieve better accuracy.  In  scheme  (\refeq{GRF22}),  $U^{(m)}$ is a vector where  the sign of the $i^{\textrm{th}}$  element of  $U^m$ indicates the class of $x_i$.  For a sample  $x_i \in \cX_{l} $ its label belongs ${\{ \pm 1 }\}$. 

Let $l_1, l_2$ denote  the number of labels in Class 1 and Class 2,  respectively with $l=1_1 + l_2$.  Then   the vector $B\in \mathbb{R}^n $ is defined   as follows
\begin{equation*}
B=\left \{
\begin{array}{ll}
 \frac{l_2}{l},   &  x_i  \in  \cX_{l}\, \textrm{and}\, y_i=1,\\
  \frac{-l_1}{l},   &  x_i  \in  \cX_{l} \,\textrm{and} \, y_i=-1,\\
  0  &        x_i\in \cX\setminus \cX_{l}.\\
  \end{array}
\right.
\end{equation*}
  Algorithm \ref{GRF1}, named  "MGRF", extends the GRF scheme by modifying the initial labels vector $Y$ and iterating over all nodes.
\begin{algorithm}
\caption{(MGRF)}\label{GRF1}
\textbf{Input:}{Matrix  $W$, initial label   $Y$, parameter $\alpha$,  tolerance $\varepsilon$.}\\
{\textbf{Output:}{ Label  $y_i$  for each point $x_i$ or  equivalently  $U\in \mathbb{R}^{n}$}}
\begin{itemize}
    \item Calculate   matrices: $D$, $S$ and $B$ 
    \item  Initialize $U^{0}= D^{-1} Y$
     \item \textbf{while} $\| U^{(m+1)} -U^{(m)}\| >\varepsilon$ 
        \textbf{do}
       \item   $ U^{(m+1)}=   \alpha S U^{(m)}  +(1-\alpha) D^{\frac{-1}{2}}B^T$
             \item  end while
             \item Assign each point  $x_{i}$   the sign of  $U^{*}_i$\\
             \item $U^{*}=\underset{m\rightarrow \infty}{Lim} U^{m}.$             
             \end{itemize}
\end{algorithm}

Next, we  explain how to improve the efficiency of  the scheme MGRF. Let $\pi$  be  the unique stationary distribution of the random walk on the graph represented by the normalized affinity matrix $S$, then $\pi$  satisfies the equation:
\[
\pi S=\pi,  
\]
$\pi$ is a row vector representing the stationary distribution.

  Consider   matrix $G$, where   all  rows of $G$ are equal and composed of   $\pi$. By subtracting  $G$  from $S$ in each iteration, we effectively bias the random walk towards the stationary distribution 
     consequently, accelerates  the convergence   towards the stationary distribution of 
     the graph.  This strategy  makes  the algorithm   robust to initial conditions and parameter choices. Furthermore, it prevents the dominance of majority classes, thus improving the overall classification performance, particularly  in imbalanced data sets.

The details of the scheme are  given in algorithm \ref{GRF2} which we call it Improved  Gaussian Random Fields (IGRF).
\begin{algorithm}
\caption{(IGRF)}\label{GRF2}
\textbf{Input:}{Matrix  $W$,     $Y$, parameters $\alpha_1, \alpha_2, \alpha_3,$ tolerance $\varepsilon$}\\
\textbf{Output:}{ Label  $y_i$  for each point $x_i$ or  equivalently  $U\in \mathbb{R}^{n}$}
\begin{itemize}
    \item Calculate   matrices:   $S$, $B$ and $\pi$
    \item  Initialize $U^{0}= D^{-1} Y$
     \item \textbf{while} $\| U^{(m+1)} -U^{(m)}\| >\varepsilon$    \textbf{do}
       \item   $ U^{(m+1)}=  (\alpha_1 S- \alpha_2 G +\alpha_3 I) U^{(m)}  +(1-\alpha_1) D^{\frac{-1}{2}} B^T$
             \item  end while
             \item Assign each point  $x_{i}$   the sign of  $U^{*}_i$\\
             \item $U^{*}=\underset{m\rightarrow \infty}{Lim} U^{m}.$
             \end{itemize}
\end{algorithm}

   \[
   \]
\subsection{Improved  Poisson Learning(IPL)}

 In this section, we aim to improve the efficiency of   Poisson Learning given by Algorithm 3 (see \cite{A8}). 
 Let $ b\in \mathbb{R}^k$ be the vector whose $i^{\textrm{th}}$ entry $b_i$
is the fraction of data points belonging to class $i$.

\begin{algorithm} 
\caption{Poisson Learning}
\textbf{Input: }{$\cX, W$,  $Y \in \mathbb{R}^{k\times l}$ (given labels), $b$, $T$:(number of iterations)  }\\
\textbf{Output:}{$U\in \mathbb{R}^{n\times k}$ predicted class labels, and classification results}
\begin{itemize}
    \item  $D  \leftarrow W \mathds{1}$
    \item $L\leftarrow  D-W$
    \item  $\overline{y}\leftarrow \frac{1}{l} Y \mathds{1} $
     \item  $B\leftarrow  [Y -\overline{y}, zeros(k, n-l)]   $
     \item   $U \leftarrow \textrm{zeros}(n,k)$
         \item \textbf{For} $m=1:T$   \textbf{do}
          \item \hskip3.0em       $ U^{(m)}=U^{(m-1)} + D^{-1} ( B^{T}- L U^{m-1})$
             \item  end for
             \item $U^m= U^m \cdot \text{diag}(b /\bar{y})$.
             \end{itemize}
\end{algorithm}

  The main step of the Poisson  Algorithm  can be rewritten as
\begin{equation} \label{Sun186}
\begin{split}
U^{(m)} & = U^{(m-1)} + D^{-1} (B^{T}- (D-W) U^{(m-1)}) \\
 & = U^{(m-1)} + D^{-1}B^{T}- U^{(m-1)} +  D^{-1}W U^{(m-1)}),\\
 & = P U^{(m-1)}+ D^{-1}B^{T}
 \end{split}
\end{equation}
where $P= D^{-1}W.$
Choosing $U^{0}=  D^{-1}B^{T}$ then (\refeq{Sun186}) implies
\[
U^{(m)}= \left( \sum_{i=0}^{m-1} P^{i}\right) D^{-1}B^{T}.
\]
Next, we  determine the limit of $P^m$ as $m$ tends to  infinity. From Markov chains theory,  for an irreducible and aperiodic Markov chain,   $P^m$   converges to a matrix where all rows are identical and represent the stationary distribution of the Markov chain.  It represents the long-term relative frequencies of being in each state of the Markov chain.   

 This stationary distribution can be computed by  finding the left eigenvector corresponding to the eigenvalue one  of the transition matrix  $P$, i.e, 
\[
\pi P=\pi.
\]
It is handy to verify  that  $\pi=(\frac{d_1}{d}, \frac{d_2}{d} \cdots ,\frac{d_n}{d}).$ 
We make ansatz that the limit of matrix  $P$  is $Q$ with matrix $Q$ being rank one matrix with rows $\pi$.
Thus
\[
\underset{m\rightarrow \infty}{Lim} P^{m}= \underset{m\rightarrow \infty}{Lim} P^{m-1} P= QP=Q.
\]

 It worths to see $P$ has  an  eigenvalue equal $one,$   means  $I-P$ is not invertible means one can not extract the fixed point for iterative scheme given by (\ref{Sun186}). This shows the sequence in (\ref{Sun186}) is not convergent.  
  
Our change  in the Poisson algorithm  is to  subtract matrix   $Q$ from $P$. Our iterative scheme   is as follows
\begin{equation*}
   U^{(m+1)}=  (P -  Q ) U^{m} +  D^{-1}B^T.
\end{equation*}
Keeping in mind the facts $ QP=Q$, and  $Q^J=Q (J=1,2,\cdots$) then it is easy to check 
\[
\underset{m\rightarrow \infty}{\text{lim}}\, (P -  Q)^m=0.
\]
Moreover, with a slightly change,   our iterative scheme   called Improved Poisson Learning  is 
\begin{equation}\label{IPL}
   U^{(m+1)}=  (P -  \alpha_1 Q ) U^{m} + \alpha_2 D^{-1}B^T,
\end{equation}
where  parameters $0<\alpha_1\le 1$ and  $\alpha_2 >0$.  
 
The scheme  updates the label matrix $U$ using a combination of transition probabilities, regularization terms, and initial label information.  It   converges faster  as it focuses entirely on the graph's intrinsic 
structure and connectivity. Furthermore, we  might  add  the term $\alpha  I $ to iterative scheme  (\ref{IPL}). 
\begin{algorithm}\label{alg:3}
\caption{(IPL)}
\textbf{Input:}{Matrix  $W$, initial label matrix $Y$, parameters $\alpha_1, \alpha_2 , \alpha_3$. }\\
\textbf{Output:}{Label  $y_i$  for each point $x_i$ or  equivalently  $U\in \mathbb{R}^{n\times k}$}
\begin{itemize}
    \item Calculate   matrices: $D$ transition matrix  $P$, $Q$
    \item  Initialize $U^{0}= D^{-1} Y$
     \item \textbf{while} not convergent    \textbf{do}
       \item   $ U^{(m+1)}=  (P - \alpha_1 Q +  \alpha_2  I)  U^{m} + \alpha_3 D^{-1}B^T$
             \item  end while
             \item Assign each point  $x_{i}$  class  $y_{i} = \underset{j}{\text{arg max}}\, U^{*}$ where\\
             \item $U^{*}=\underset{m\rightarrow \infty}{Lim} U^{m}$
             \end{itemize}
\end{algorithm}
Adding $\alpha_2 I $  to the transition matrix $P$  effectively increases the self-transition probabilities of nodes in the graph by $\alpha_2$. This can be interpreted as giving more weight to the existing labels in the updating process.
It may help stabilize the iterative process by ensuring that each node retains some influence from its current label in the next iteration. The addition of  $\alpha_2 I $ can also help prevent numerical instability issues, particularly since 
$P$ and  $Q$ have eigenvalues close to one.

\section{Experimental results}

In this section,  we  compare the proposed  algorithms with some existing  semi-supervised learning algorithms.   
For imbalanced data, we use  evaluation metrics that are robust, such as precision, recall, and  ${\rm F}_{1}$-score. These metrics provide a more comprehensive assessment of model performance on imbalanced data compared to accuracy.

For simplicity reason, we name   the minor class as positive class interchangeably.   Precision measures the accuracy of the positive predictions. It is the ratio of true positive predictions to the total predicted positives.
Recall measures the ability to capture all positive samples. It is the ratio of true positive predictions to the total actual positives. $F_1$  Score is the harmonic mean of precision and recall. It balances the two metrics, especially useful in the case of imbalanced data sets.  The term of $F_1$ score  for each class is defined as follows: 
\begin{equation}
\label{F1measure}
    {\rm F}_1(j) = 2 \times \dfrac{{\rm{Recall}}(j) \times {\rm{Precision}}(j)}{{\rm{Recall}}(j) + {\rm{Precision}}(j)} \quad j= 1,2.
\end{equation}
Accuracy measures the overall correctness of the model. It is the ratio of all correct predictions to the total number of samples.
 
\begin{example}
 
We consider balanced  \textit{Two-Moon} pattern, we generate a set of 1000  points, with noise level $0.15$. Increasing noise will make classes more overlapping.   In Algorithm  IPL,  the   parameters are set  as  $\alpha_1= 0.001, \alpha_2=0.02, \alpha_3=1$. 
\begin{figure}[h]
  \includegraphics[scale=0.38]{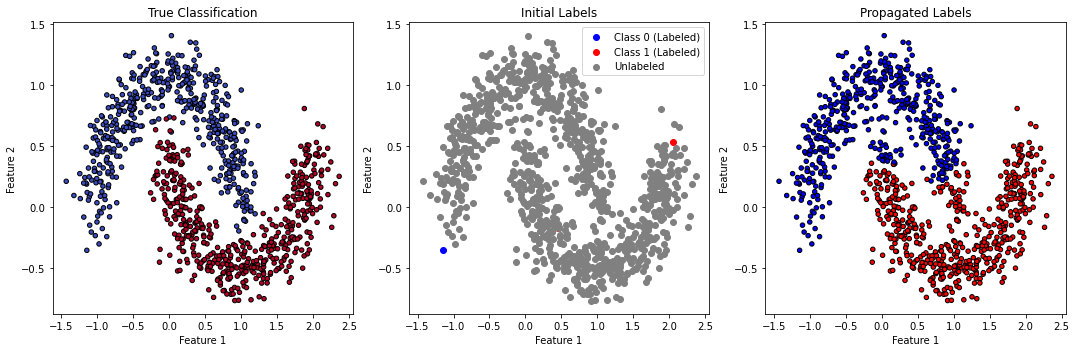}
\caption{The classification on  Two-Moon.}
\label{fig:1}       
\end{figure}
\vspace{.5cm}

In Table \ref{tab:single_row_table}, we present a comparison between Algorithms IGRF and IPL, utilizing the two moons dataset consisting of $n=1000$ points with a noise level set at $0.15$.

  For Algorithm IGRF, we construct the affinity matrix using the RBF (Gaussian) kernel, which offers a smoother affinity measure than the $k$-nearest neighbors graph. The  parameters of IGRF are set  as $ \alpha_1=.99,  \alpha_2 = 0.005,     \alpha_3 = 0.05.$  
  
 In the IPL algorithm, we employ a grid search over hyperparameters to determine the optimal parameters for various numbers of initial labels. However, this process is time-consuming. The table displays the average accuracy for each number of initial labels per class. 

  \begin{table}[htbp]
    \centering
    \caption{Two moon, balanced  case}
    \begin{tabular}{|c|c|c|c|c|c|}
        \hline
        \multicolumn{6}{|c|}{\textbf{Average overall accuracy   over 100 trials for two moon}} \\
        \hline
         number of labels per class   &  1 &  2 &  3 &  4 &  5  \\
        \hline
           IGRF     &  \textbf{94.64}  &  \textbf{97.15}  & \textbf{98.05}  & \textbf{98.348}  &  \textbf{98.53}       \\
        \hline
          IPL    &   86.30  & 89.55  & 91.86  & 92.12         &  93.04       \\
        \hline
        Poisson    &   83.23  & 87.985  & 90.406  & 91.953         &  92.70       \\
        \hline
    \end{tabular}
    \label{tab:single_row_table}
\end{table}
\end{example}
 
\begin{example}
Next, we generate  imbalance two half moons with 950 samples from Majority class and 50 samples from  Minority. We implement the IGRF scheme with two labeled points per classes.  
The  average of accuracy over 100 runs is $ 96.765$.  Also we compute the average of the Confusion Matrix over 100 runs.
\begin{displaymath}
\left(\begin{array}{cc}
30.27 &    19.73\\
 11.62  &  938.38\\
\end{array}\right).
\end{displaymath}

\begin{figure}[h]
\includegraphics[scale=0.4]{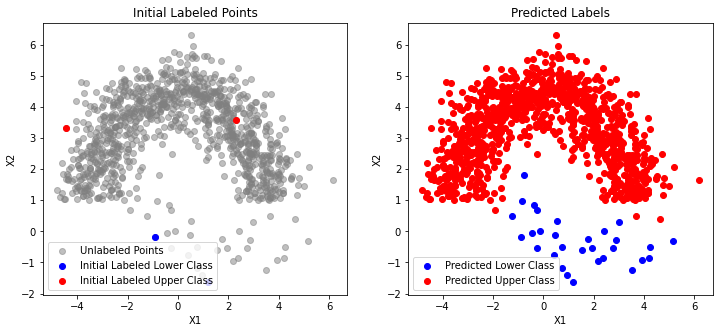}
\caption{The classification on  Imbalanced  Two-Moons. }
\label{fig:2}       
\end{figure}
\end{example}

\begin{example}
   In this example, we implement Algorithm 1 (MGRF) and analyze a scenario where the data points are distributed across two circles. The configuration is set as follows: the  major class contains  1,000 samples,while the minor class contains 100; noise level is set at 0.1; the maximum number of iterations is 1,500; the tolerance is set at $=10^{-8}$, $\alpha=.985$  and there are 3 labeled samples per class. Refer to Figure \ref{fig:3} for more details. The average confusion matrix, calculated over 100 trials, is presented below. Figure \ref{fig:3} also displays the initial labeled points and the predictions made by the MGRF algorithm.
  \begin{displaymath}
\left(\begin{array}{cc}
919.72 & 80.28\\
 25.47 &  74.53\\
 \end{array}\right).
\end{displaymath}

 \begin{figure}[h]
\includegraphics[scale=0.4]{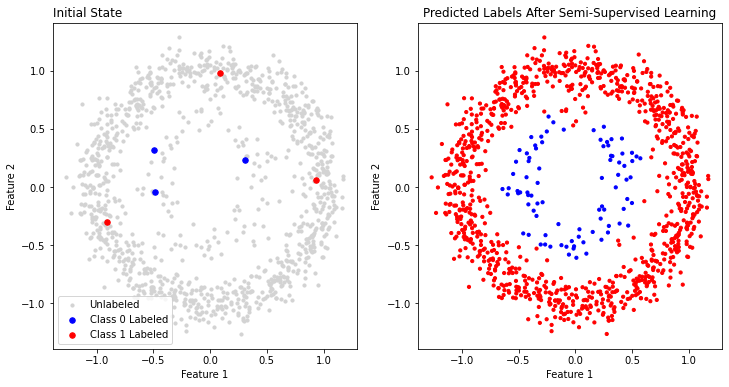}
\caption{The classification on   (\textit{two-circles} ). }
\label{fig:3}       
\end{figure} 
\end{example}

\begin{example}
   Table \ref{Table4}  describes the  accuracy $(\%) $ 
of    IPL algorithm compared to  Poisson and Segregation algorithms,  on Cifar-10 data set  with 3 classes and 4500 nodes (1500 nodes per class)  with several labels per  class. The results are   averages over 100 trials, which demonstrates   superior performance of our scheme IPL for  different label rates. 
   \begin{table}[htbp]
    \centering
    \caption{Cifar-10 data set}
    \begin{tabular}{|c|c|c|c|c|c|c|c|}
        \hline
        \multicolumn{8}{|c|}{\textbf{Average accuracy   over 30 trials for 3 classes on Cifar-10 dataset}} \\
        \hline
         number of labels    &  2 &  3 &  4 &  5 &  10  &   20 &  40 \\
        \hline
       Poisson        &   37.5  &  39.5   &  40.3  &   41.4     &   44.10    &  48.7    &  50.3      \\
        \hline
        Segregation   & 34.9    &  35.6    &  36.2  &  38.6     &    42.4   &  45.3    &  47.7       \\
        \hline
        IPL           & \textbf{45.226}  & \textbf{46.546 }  & \textbf{47.382} &  \textbf{48.341}   &  \textbf{53.152}  & \textbf{54.966}  &  \textbf{57.031}     \\
        \hline
    \end{tabular}
    \label{Table4}
\end{table}
Dimensionality reduction techniques like PCA or feature transformations provide better feature separation for the proposed algorithms, however, we did not  involve PCA in algorithm. We just used PCA, in Figure \ref{fig:13}   to plot  the classification for 3 classes on the Cifar-10 data set.
  \begin{figure}[h]
\includegraphics[scale=0.4]{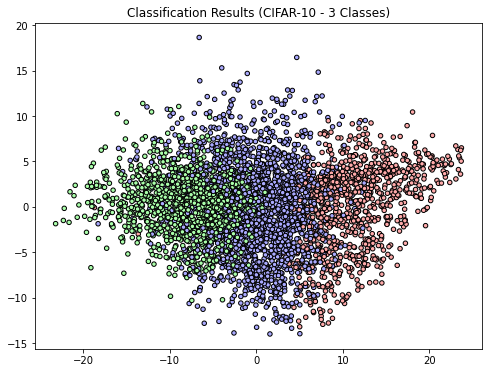}
\caption{The classification of   \textit{Cifar-10}. }
\label{fig:13}       
\end{figure} 
\end{example}

\begin{example}
We tested  our proposed method on 10 KEEL imbalanced benchmark data sets which contains  different levels of imbalance in data and  different sample sizes \cite{A} as summarized in Table  \ref{Table:datasets}.  In Table. \ref{Table:datasets} the column denoted by  $n$ stands for the total number of samples, $p$ is  the number of features, and    IR indicates the ratio  between majority and Minority class samples in each data set. 
{\scriptsize{
\begin{table}[!htbp]
  \centering\footnotesize
  \caption{Information of selected imbalanced benchmark datasets}
    \begin{tabular}{l|cccr}    
\hline
\textbf{Dataset} & \textit{IR} & {$p$} & $n$ & \%Minority\\
\hline
    \textbf{abalone9-18}	&	16.40	&	8	&	731	&	5.74	\\
    \textbf{appendicitis}	&	4.04	&	7	&	106	&	19.81	\\
    \textbf{ecoli2}	&	5.46	&	7	&	336	&	15.47	\\
    \textbf{hypo}	&	15.49	&	6	&	2012	&	6.06	\\
    \textbf{new-thyroid1}	&	5.14	&	5	&	215	&	16.27	\\
    \textbf{shuttle-c0-vs-c4}	&	13.86	&	9	&	1829	&	6.72	\\
    \textbf{sick}	&	11.61	&	6	&	2751	&	7.92	\\
    \textbf{vowel0}	&	9.98	&	13	&	988	&	9.1	\\
    \textbf{yeast3}	&	8.10	&	8	&	1484	&	10.98	\\
    \hline
    \end{tabular}%
    \label{Table:datasets}
\end{table}
}}
To evaluate our Algorithms,  we  use the  metrics   \textit{${\rm F}_1$}-Score,   Recall,  Accuracy and Precision for each classes. To ensure consistency  for all experiments, for each benchmark, first, the data set is shuffled. Subsequently, 1 percent of the samples are randomly chosen in accordance with the dataset's IR   as the labeled samples. This process is independently repeated 100 times, then   the averages of the previously mentioned metrics are computed. I should mention that for data set "shuttle-c0-vs-c4" we choose 4 labels per class which is $0.43\%$ of data. We did not compare with existing schemas since most schemes with only  $0.43\%$ of labeled samples per class give poor results.

     Table \ref{4} displays the comparison  between  performance of our proposed method, with Poisson Learning.   For each data set, the second column
       indicates the number of samples randomly selected from each class as labeled data.

 {\scriptsize{
 \begin{table}[!htbp]
  \centering\footnotesize
  \caption{Information of selected imbalanced benchmark datasets}
  \begin{tabular}{|c|c|c|c|c|c|c|c|c|c|c|}

    \hline

     Names &  \multicolumn{7}{|c|}{IGRF} \\ 

    \hline

    Dataset &  Accuracy & F1 min & F1 maj & Recall min & Recall maj & Precision min & Precision maj\\ %
    \hline

    yeast3   &   $ .864$ & $.524$ & $ .919$ & $.693$ & $.882$ & $.421$ & $.959$\\

    \hline

    appendicitis &    $.758$ & $.529$ & $.837$ & $.705$ & $.771$ & $.423$ & $.917$\\

    \hline

    abalone9-18 &    $.9519$ & $.7356$ & $.9749$ & $.7380$ & $.9941$ & $.7333$ & $.9565$\\

    \hline

    ecoli2 &    $.875$ & $.694$ & $.922$ & $.912$ & $.869$ & $.561$ & $.982$\\

    \hline
    hypo &   $.9024$ & $.7257$ & $.5657$ & $.82352$ & $.9169$ & $.6488$ & $.9407$\\

    \hline

    new-tyroid1 &   $.877$ & $.425$ & $.931$ & $.278$ & $.994$ & $.898$ & $.876$\\

%

    \hline

    \textbf{shuttle-c0-vs-c4} & \textbf{.9737} &  $\textbf{.757}$ & \textbf{.986} &\textbf{.6097} & \textbf{ .986} &   \textbf{1}  &\textbf{.970}  \\

    \hline

    sick &    $.8146$ & $.4044$ & $.8196$ & $.7833$ & $.8196$ & $.2725$ & $.8197$\\

    \hline

    vowel0 &    $.725$ & $.172$ & $.835$ & $.317$ & $.766$ & $.118$ & $.919$\\

    \hline
    \end{tabular}
 \label{4}
\end{table}
}}
\end{example}
%
%
%
\section{Conclusion}
In this work,  we present several schemes  to  enhance  the efficacy of classification algorithms when dealing with large datasets with limited labeled data.   The method leverage  the efficiency of Poisson  and  Gaussian Random Fields methods  while maintains   simplicity and convergence.  

The Modified Gaussian Random Fields (MGRF) and IGRF algorithms optimize  label propagation across a graph by adjusting the initial label vector and iteratively refining the classification process. This approach demonstrates a substantial improvement.
 
The Improved Poisson Learning (IPL) algorithm,   addresses the limitations of  Poisson Learning by incorporating a subtraction of the matrix from the transition matrix, enhancing the stability and convergence of the label propagation process. This adjustment allows the algorithm to perform more effectively the case that  data is  imbalanced, ensuring that minority classes are   represented and classified.

\renewcommand{\refname}{REFERENCES }

\end{document}